\documentclass{IOS-Book-Article}

\usepackage{mathptmx}
\usepackage{amsmath}
\usepackage{graphicx}
\usepackage{subcaption}
\usepackage{bm}
\usepackage{xcolor}
\usepackage{hyperref}
\usepackage{enumitem}
\usepackage{colortbl}
\usepackage{algorithm}
\usepackage{algpseudocode}
\usepackage{amsfonts}

\begin{document}
\begin{frontmatter}              

\title{Limits of Generative Pre-Training in Structured EMR Trajectories with Irregular Sampling}
\runningtitle{Limits of Generative Pre-training for Irregular EMR Trajectories}

\author[A]{\fnms{Nicholas I-Hsien Kuo}%
\thanks{Corresponding Author: Nicholas I-Hsien Kuo; E-mail: n.kuo@unsw.edu.au\newline}},
\author[A]{\fnms{Blanca Gallego}}
and
\author[A]{\fnms{Louisa R Jorm}}

\runningauthor{Kuo et al.}
\address[A]{ Centre for Big Data Research in Health, Faculty of Medicine,\\ 
University of New South Wales, Sydney, NSW, Australia}
ORCiD ID: Nicholas I-Hsien Kuo https://orcid.org/0000-0001-8749-7280

\begin{abstract}
Foundation models refer to architectures trained on vast datasets using autoregressive pre-training from natural language processing to capture intricate patterns and motifs. They were originally developed to transfer such learned knowledge to downstream predictive tasks. Recently, however, some studies repurpose these learned representations for phenotype discovery without rigorous validation, risking superficially realistic but clinically incoherent embeddings. To test this mismatch, we trained two autoregressive models -- a sequence-to-sequence LSTM and a reduced Transformer -- on longitudinal ART for HIV and Acute Hypotension datasets. Controlled irregularity was added during training via random inter-visit gaps, while test sequences stayed complete. Patient-trajectory synthesis evaluated distributional and correlational fidelity. Both reproduced feature distributions but failed to preserve cross-feature structure -- showing that generative pre-training yields local realism but limited clinical coherence. These results highlight the need for domain-specific evaluation and support trajectory synthesis as a practical probe before fine-tuning or deployment. 

\end{abstract}

\end{frontmatter}

\thispagestyle{empty}
\pagestyle{empty}

\section{Introduction}

Foundation neural network models learn general-purpose representations from large datasets~\cite{bommasani2021opportunities}, enabling fine-tuning for tasks such as readmission or treatment response prediction~\cite{khan2025comprehensive}. They also support representation learning for phenotype discovery, to uncover latent biomedical structure~\cite{khan2025comprehensive}. Most use autoregressive generative pre-training~\cite{radford2019gpt2}, predicting the next element from prior context -- the basis of large language models (LLMs). Recently, models such as ETHOS~\cite{renc2024zero} applied this to structured electronic medical records (EMRs). Yet EMRs are not text: sequential coherence rarely holds as patient trajectories reflect informative missingness from clinical decision-making~\cite{li2021imputation}.

This study examines whether the autoregressive generative pre-training paradigm remains meaningful for clinical time series and how its “goodness-of-fit” should be evaluated. Using two longitudinal datasets (antiretroviral therapy [ART] for HIV and acute hypotension), we train a sequence-to-sequence LSTM~\cite{sutskever2014seq2seq} and a reduced-scale ETHOS-style Transformer under controlled temporal irregularity. We assess each model’s ability to encode past patient profiles and extrapolate forward by measuring distributional and correlational fidelity. Probing embeddings is crucial when such models are used for phenotype discovery, ensuring learned structure reflects true clinical relations. Results show that generated trajectories preserve feature distributions but lose cross-feature coherence, highlighting the need for rigorous, domain-specific evaluation before deploying natural language processing-style foundation models in healthcare.

\section{Methods}
\subsection{The Datasets}\label{Sec:Features}
\begin{table}[t]
\centering
\caption{Baseline characteristics for the ART for HIV and the acute hypotension datasets.}
\label{tab:baseline}
\begin{tabular}{p{2.4cm} p{2.8cm} || p{2.65cm} p{2.7cm}}

\multicolumn{2}{l}{\textbf{ART for HIV}} & \multicolumn{2}{l}{\textbf{Acute Hypotension}}\\
\hline
\hline
\rowcolor{red!10}\multicolumn{4}{l}{\textbf{Numeric features (median [interquartile range])}} \\
\hline
\hline

\rowcolor{gray!10}
Viral load\newline(VL, copies/mL)   & 
38.78 [10.49, 778.27] &
Mean Arterial Pressure\newline(MAP, mmHg) &
65.34 [59.30, 71.19]
\\

CD4 count\newline(cells/µL) & 
466.40 [272.85, 859.44] &
Urine\newline (mL)&
106.21 [68.92, 164.23]
\\

\rowcolor{gray!10}
& &
Lactate\newline(mmol/L) &
1.50 [1.29, 1.80]
\\

\hline
\hline
\rowcolor{blue!10}\multicolumn{4}{l}{\textbf{Categorical features (percentage share \% of each unique level)}} \\
\hline
\hline
\rowcolor{gray!10}
Base combo & 
FTC + TDF: 48.2 &
Vasopressors& 
0: 84.14\\
\rowcolor{gray!10}
& 3TC + ABC: 29.2 & (mcg/kg/min) & 
(0, 8.4): 8.34\%\\
\rowcolor{gray!10}
& FTC + TAF: 2.5 &  & 
[8.4,20.28): 3.68\%\\
\rowcolor{gray!10}
&  
DRV + FTC + TDF: 13.9 & &
$\ge$20.28: 3.83\%\\
\rowcolor{gray!10}
& 
FTC + RTVB + TDF: 4.8 & 
&\\
\rowcolor{gray!10}
& Other: 1.4 &  & \\
Comp. INI      &
DTG: 22.6  & Fluid Boluses & [0,250): 97.32\% \\
& RAL: 4.3 & (mL) & [250,500): 0.28\% \\
& EVG: 6.9 & & [500,1000): 1.46\% \\
& Not applied: 66.2 & & $\ge$1000: 0.94\%\\
\rowcolor{gray!10}
Extra PI        &  
DRV: 6.4 \hspace{2mm}RTVB: 10.0 &  & \\
\rowcolor{gray!10}
& LPV: 0.1 \hspace{2.375mm}RTV: 2.6 & & \\
\rowcolor{gray!10}
& ATV: 4.7 & & \\
\rowcolor{gray!10}
& Not applied: 76.2 & & \\
\hline
\multicolumn{4}{l}{FTC: Emtricitabine; TDF:Tenofovir disoproxil fumarate; 3TC: Lamivudine; ABC: Abacavir;}\\
\multicolumn{4}{l}{TAF: Tenofovir alafenamide; DRV: Darunavir; RTV(B): Ritonavir (boosted); DTG: Dolutegravir;}\\
\multicolumn{4}{l}{RAL: Raltegravir; EVG: Elvitegravir; LPV: Lopinavir; ATV: Atazanavir}
\end{tabular}
\end{table}

We considered two longitudinal datasets: ART for HIV and Acute Hypotension, summarised in Table~\ref{tab:baseline}. The former captures antiretroviral therapy responses using five key features -- \textbf{viral load}, \textbf{CD4 count}, \textbf{Base Combo} (ART backbone), \textbf{Comp. INI} (integrase inhibitor use), and \textbf{Extra PI} (protease inhibitor use) -- while the latter describes management in ICU with \textbf{MAP}, \textbf{Urine output}, \textbf{Lactate}, \textbf{Vasopressors}, and \textbf{Fluid Boluses}. The ART for HIV dataset, openly accessible in~\cite{nicholas2023generating} based on EUResist~\cite{zazzi2012predicting}, contains 8,916 patients sampled monthly; the Acute Hypotension dataset, derived from the MIMIC-III database~\cite{johnson2020mimic,gottesman2020interpretable} and available under credentialed access, includes 3,910 patients sampled hourly. Both datasets were standardised into fixed-length trajectories -- 60 months and 48 hours, respectively -- with no missing values to ensure consistent model inputs.

\subsection{Dual Split Strategy: Controlled Irregular Sampling and \texorpdfstring{$\Delta t$}{Δt} Encoding}

\begin{figure}[h]
\centering
\includegraphics[width=0.6\linewidth]{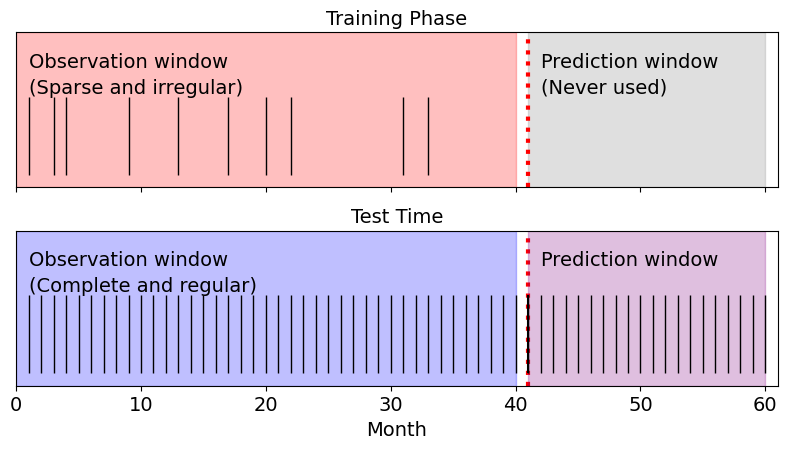}
\caption{Demonstrating the dual-split design using the ART for HIV dataset.}
\label{fig:Irregularity}
\end{figure}

Each dataset was divided using a \textbf{dual-split design} to evaluate cross-patient generalisation and temporal extrapolation. 
An 80--20 train--test split prevented information leakage, and a 2:1 temporal split (40:20 months for ART for HIV; 32:16 hours for acute hypotension) separated trajectories into an observation and a prediction window (Figure~\ref{fig:Irregularity}). 
During training (\textcolor{pink}{pink}), models were fitted autoregressively on the observation window, where visits were subsampled to simulate irregular follow-up. 
The prediction window was unseen during training and used to assess the goodness of fit via synthesis quality of future trajectories (\textcolor{purple}{purple}). 
Controlled irregularity was introduced only during training by sampling random inter-visit gaps \( g \sim \mathrm{Uniform}(1,G_{\max}) \), with \( G_{\max}=10 \) and \( 35 \) months for ART for HIV and proportionally equivalent 8- and 28-hour gaps for acute hypotension. 
The elapsed interval between retained visits (\( \Delta t \)) was included as an input feature to inform the model of variable temporal spacing.

\subsection{Autoregressive Models}
\label{sec:model}

We evaluated two architectures: an LSTM encoder–decoder~\cite{sutskever2014seq2seq} (1 layer, 64 hidden units) and a reduced ETHOS Transformer (ETHOS-lite, 2 layers)~\cite{renc2024zero}. The original ETHOS used a GPT-2–style decoder~\cite{radford2019gpt2} trained on large-scale MIMIC-IV data; here, we down-scaled the model to mitigate overfitting on our smaller datasets. Both models were trained for 10 epochs with Adam~\cite{kingma2014adam} (learning rate $10^{-3}$) to minimise feature reconstruction loss over the observation window. All code will be publicly released upon acceptance.

\section{Results}
When extrapolating beyond the observation window (months 41–60 for ART for HIV; hours 33–48 for Acute Hypotension), both models reproduced test-set marginals even 

\newpage
\begin{figure}[h]
  \centering
    \includegraphics[width=0.965\linewidth]{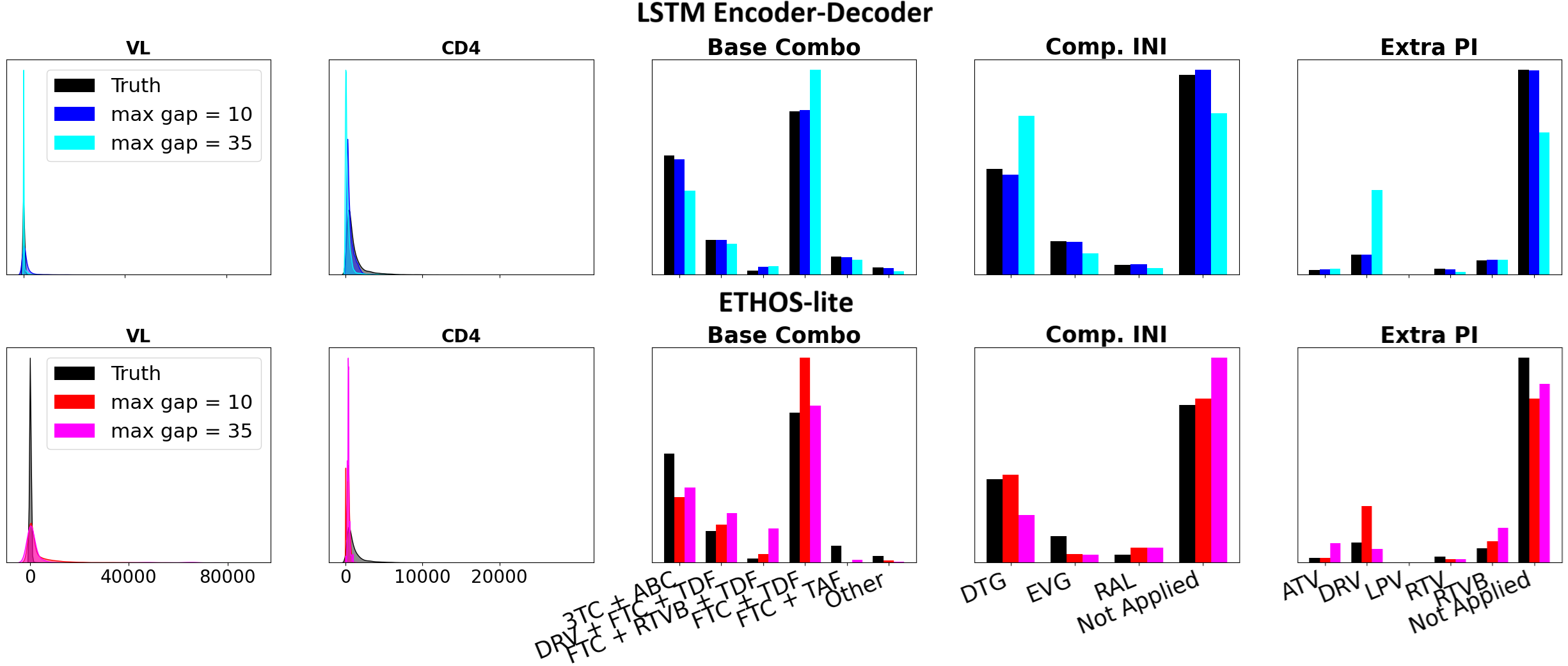}
    \includegraphics[width=0.965\linewidth]{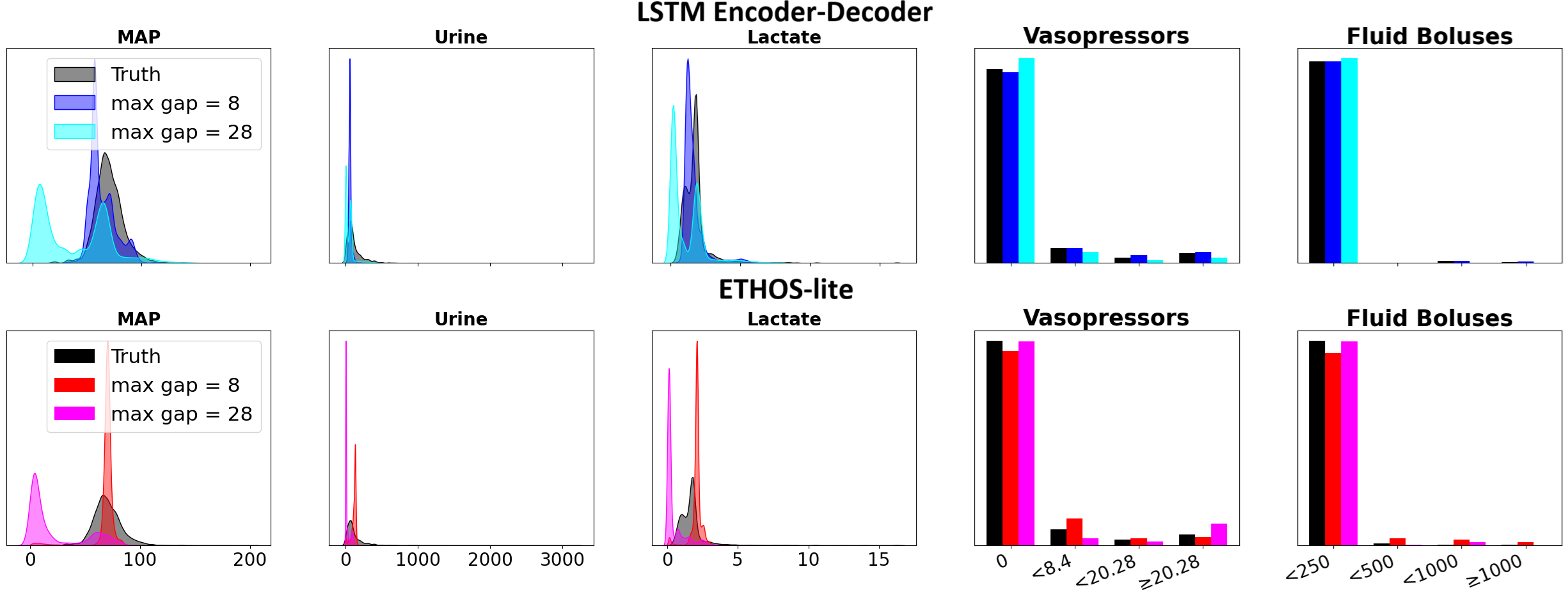}
  \caption{Distributional fidelity: ART for HIV (rows 1–2) and Acute Hypotension (rows 3–4).}
  \label{fig:Results1}
\end{figure}
\begin{figure}[h]
  \centering
  \includegraphics[width=0.965\linewidth]{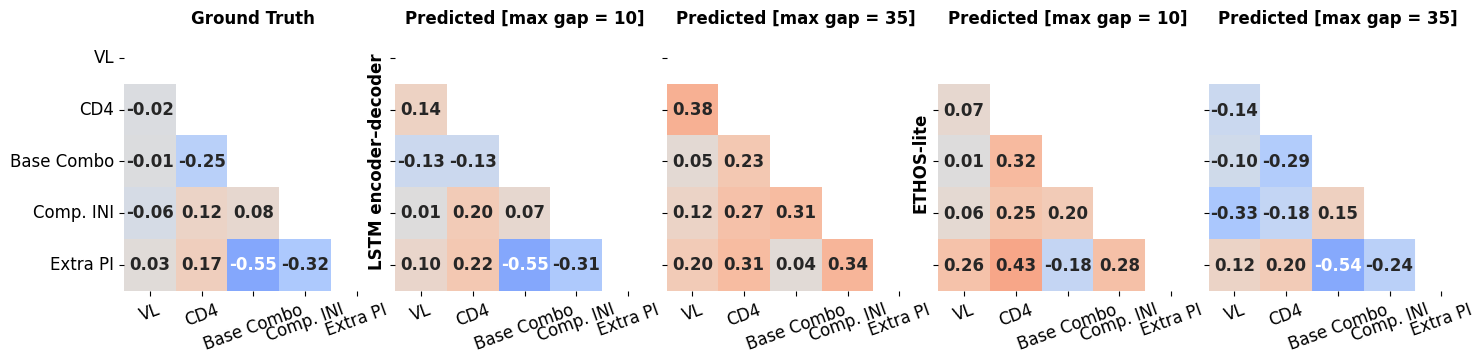}
  \includegraphics[width=0.965\linewidth]{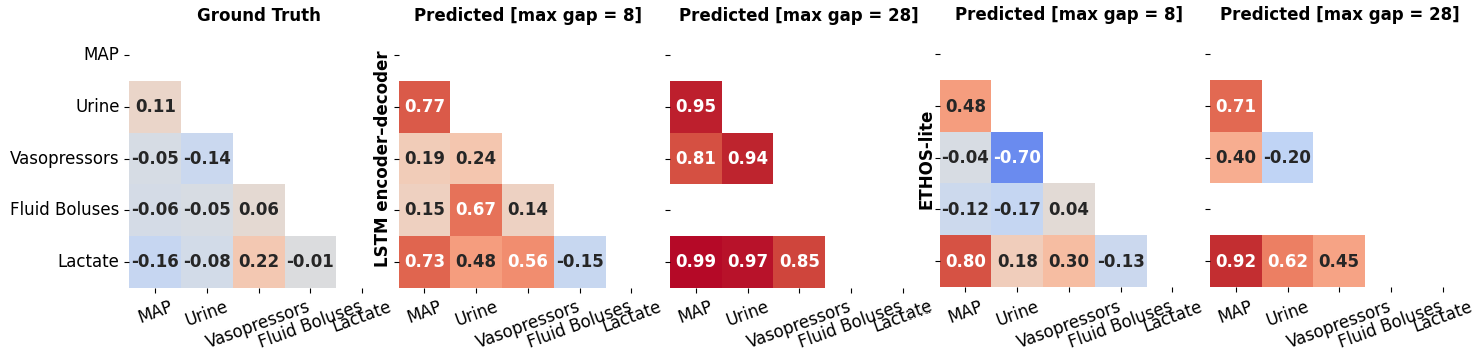}

  \caption{Correlational fidelity: ART for HIV (row 1) and Acute Hypotension (row 2).}
  \label{fig:Results2}
\end{figure}

\newpage
\hspace*{-6mm}under severe irregularity (35 months / 28 hours), showing strong distributional fidelity (Figure~\ref{fig:Results1}). However, correlational fidelity was inconsistent (Figure~\ref{fig:Results2}), indicating that both models failed to capture underlying clinical dependencies. Some tiles are missing as certain categorical variables had only one level, preventing correlation computation.

\section{Discussion}
Across two longitudinal cohorts (ART for HIV and acute hypotension) with irregular sampling to mimic informative missingness, autoregressive pre-training preserved distributional realism but failed to maintain correlational structure. The LSTM encoder-decoder retained limited dependencies under moderate gaps, while ETHOS-lite produced inconsistent results. These findings suggest that autoregressive pre-training does not transfer cleanly from natural language to structured EMRs, yielding realistic but clinically incoherent trajectories. Accordingly, such models may offer a foundation for downstream prediction, but their use for knowledge discovery or patient phenotyping remains premature. Patient-trajectory synthesis provides a complementary probe, exposing weaknesses in representation fidelity and explainability.

\section{Conclusion}
Autoregressive foundation models fail to preserve cross-feature coherence under irregular sampling, highlighting the need for domain-expert stress-testing before deployment.


\end{document}